\documentclass{article}
\usepackage{spconf,amsmath,graphicx,hyperref}
\usepackage{amssymb}
\usepackage{mathptmx}
\usepackage{multirow}
\usepackage{multicol}
\usepackage{float}
\usepackage{balance}
\usepackage{multicol}
\usepackage[table]{xcolor} 
\usepackage{array}  
\usepackage[utf8]{inputenc}

\title{U-MAN: U-Net with Multi-scale Adaptive KAN Network for Medical Image Segmentation}
\name{Bohan Huang$^{1}$ \qquad Qianyun Bao$^{2}$ \qquad Haoyuan Ma$^{3}$\thanks{© 2025 IEEE. Personal use of this material is permitted. Permission from IEEE must be obtained for all other uses, in any current or future media, including reprinting/republishing this material for advertising or promotional purposes, creating new collective works, for resale or redistribution to servers or lists, or reuse of any copyrighted component of this work in other works.}}

\address{$^{1}$Nanjing University of Posts and Telecommunications, Nanjing, China\\
    $^{2}$Nanjing Normal University of Special Education, Nanjing, China\\
    $^{3}$Soochow University, Suzhou, China}
\begin{document}
\ninept
\maketitle
\begin{abstract}
Medical image segmentation faces significant challenges in preserving fine-grained details and precise boundaries due to complex anatomical structures and pathological regions. These challenges primarily stem from two key limitations of conventional U-Net architectures: (1) their simple skip connections ignore the encoder-decoder semantic gap between various features, and (2) they lack the capability for multi-scale feature extraction in deep layers. To address these challenges, we propose the U-Net with Multi-scale Adaptive KAN (U-MAN), a novel architecture that enhances the emerging Kolmogorov-Arnold Network (KAN) with two specialized modules: Progressive Attention-Guided Feature Fusion (PAGF) and the Multi-scale Adaptive KAN (MAN). Our PAGF module replaces the simple skip connection, using attention to fuse features from the encoder and decoder. The MAN module enables the network to adaptively process features at multiple scales, improving its ability to segment objects of various sizes. Experiments on three public datasets (BUSI, GLAS, and CVC) show that U-MAN outperforms state-of-the-art methods, particularly in defining accurate boundaries and preserving fine details. 
\end{abstract}
\begin{keywords}
Medical Image Segmentation, Kolmogorov-Arnold Networks (KAN), U-Net, Deep learning, Feature Fusion
\end{keywords}

\section{Introduction}
\label{sec:intro}
Medical image segmentation is indispensable for modern clinical workflows \cite{shen2017deep}, providing quantitative insights for diagnosis \cite{litjens2017survey}, treatment planning \cite{isensee2021nnu}, and disease monitoring \cite{noble2006ultrasound}. However, the reliability of segmentation is often compromised by inherent challenges such as low contrast, noise, irregular boundaries, and high inter-patient variability \cite{hesamian2019deep}.

Deep learning approaches have achieved remarkable success in medical image segmentation. The U-Net architecture laid the foundation for this field with its efficient encoder-decoder structure and skip connections \cite{ronneberger2015u}; However, its reliance on standard convolutions limits its ability to model complex non-linear relationships and makes it prone to overfitting on small datasets. Attention U-Net integrated attention gates to focus on salient regions \cite{oktay2018attention}, but often at the cost of higher computational demand and potential neglect of critical contextual information.
A paradigm shift occurred with the introduction of TransUNet \cite{chen2021transunet}, which leveraged Transformers for powerful global feature modeling and achieved substantial accuracy improvements \cite{shamshad2023transformers}. However, its performance depends on large-scale training data and is accompanied by marked increases in model complexity and computational cost. 
A common limitation across existing research is the lack of an architecture that can effectively model non-linear features without inheriting the computational burdens or data dependencies of its predecessors.

Although U-KAN integrates KAN into U-Net, it still has limitations. It retains conventional, simple skip connections that perform direct feature fusion, failing to bridge the semantic gap between shallow (detail-rich) and deep (context-rich) features. Additionally, standard KAN blocks, despite being powerful function approximators, lack adaptive multi-scale feature processing mechanisms required in computer vision \cite{zhao2017pyramid}.

To address these limitations, we propose a novel architecture U-Net with Multi-scale Adaptive KAN (U-MAN). First, to resolve the semantic gap caused by naive skip connections, we introduce the Progressive Attention-Guided Feature Fusion (PAGF) module. PAGF replaces direct feature merging with a sophisticated attention mechanism \cite{vaswani2017attention} that intelligently filters and reweights features before fusing them with the decoder pathway. Second, to empower the network with adaptive multi-scale processing capabilities, we design the Multi-scale Adaptive KAN (MAN) module. MAN enhances the standard KAN blocks with a dual-branch attention architecture, enabling the model to dynamically capture contextual information at various scales and improve its performance on objects of different sizes, which is crucial for handling anatomical structures with varying dimensions in medical imaging \cite{isensee2021nnu}.

The main contributions of this work are threefold:
\begin{itemize}
    \item We propose a novel architecture called U-MAN that systematically addresses the semantic gap in KAN-based U-Net frameworks. The PAGF modle alleviates semantic gaps among feature maps from different layers.
    \item We design the MAN module to capture deep multi-scale feature maps, enabling the network to adaptively capture feature at different scales.
    \item We conduct extensive experiments on three public benchmark datasets (BUSI, GLAS, and CVC-ClinicDB). Our results validate that U-MAN consistently outperforms existing state-of-the-art methods, demonstrating its effectiveness and generalization capabilities across different medical imaging modalities.
\end{itemize}

\section{Method}
\label{sec:format}

\subsection{Problem Formulation and Network Overview}

\subsubsection{Problem Formulation}
Given a medical image $\mathbf{I} \in \mathbb{R}^{H \times W \times C}$, we define $H$, $W$, and $C$ as height, width, and channels, respectively. Our objective is to learn a mapping function $f: \mathbf{I} \rightarrow \mathbf{Y}$ that produces an accurate segmentation mask $\mathbf{Y} \in \mathbb{R}^{H \times W \times K}$ \cite{long2015fully}. Here, $K$ represents the number of segmentation classes. We let $\boldsymbol{\Theta}$ denote the learnable parameters of our model. The optimization objective can be formulated as:

\begin{equation}
\boldsymbol{\Theta}^* = \arg\min_{\boldsymbol{\Theta}} \mathcal{L}(f(\mathbf{I}; \boldsymbol{\Theta}), \mathbf{Y}_{gt})
\end{equation}
where $\mathbf{Y}_{gt}$ represents the ground truth segmentation mask and $\mathcal{L}$ denotes the loss function \cite{milletari2016v}.

\subsubsection{Network Overview}
\begin{figure}[htbp]
    \centering
    \includegraphics[width=\columnwidth]{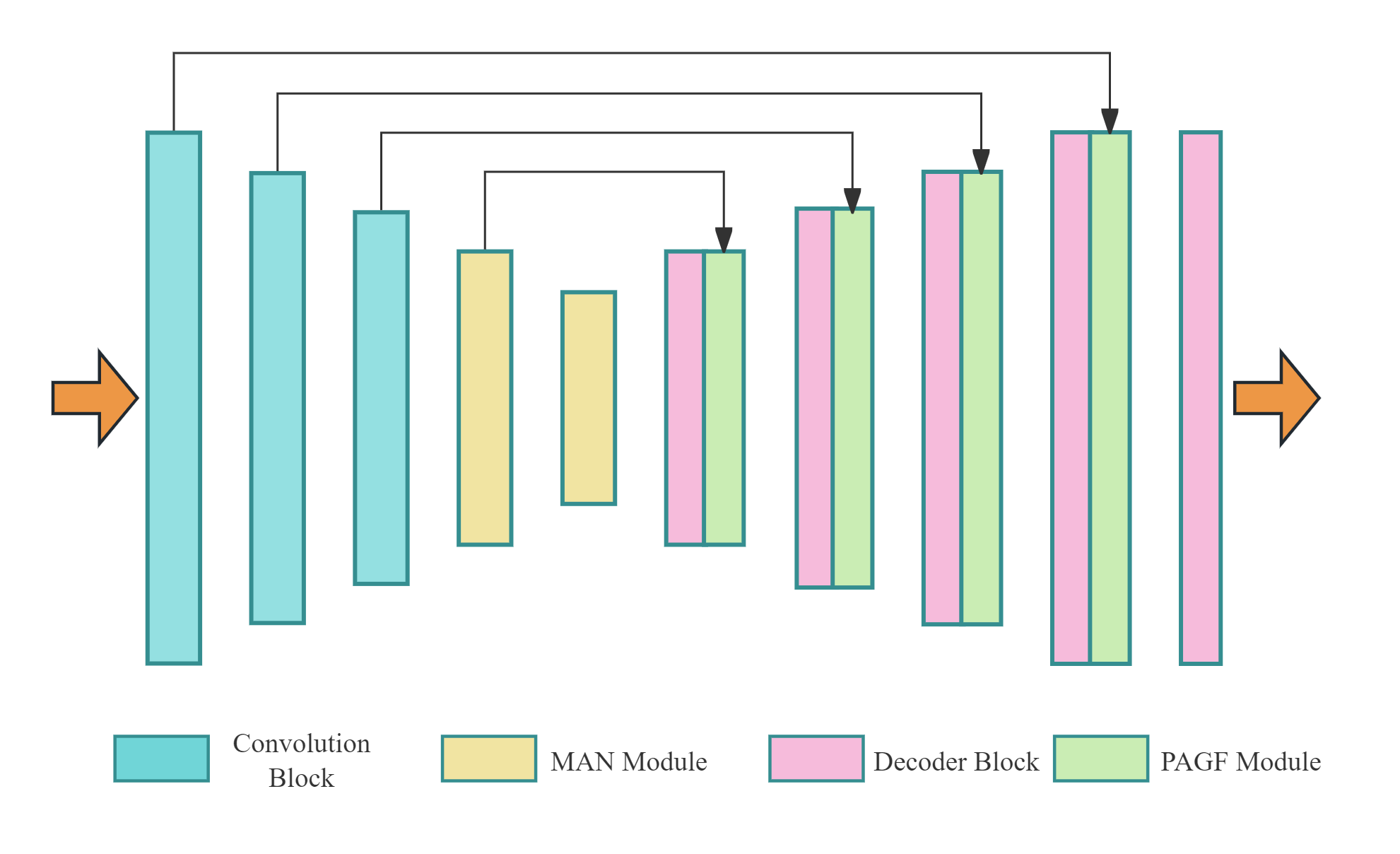}
    \caption{Overall architecture of U-MAN. The encoder contains Convolution Blocks for shallow feature extraction andMAN modules perform adaptive multi-scale processing in the deeper layers. Critically, each skip connection is enhanced by a PAGF Module to intelligently fuse features from the encoder and decoder paths.}
    \label{fig:overview}
\end{figure}

As illustrated in Figure \ref{fig:overview}, U-MAN follows a hybrid encoder-decoder architecture \cite{ronneberger2015u}. The architecture systematically integrates convolutional stages with multi-scale attention processing \cite{vaswani2017attention}. The encoder consists of three convolutional stages followed by two MAN stages. It generates multi-scale feature representations with embedding dimensions $\{32, 64, 256, 320, 512\}$ at different resolution levels \cite{he2016deep}. The decoder pathway reconstructs segmentation masks through progressive upsampling \cite{long2015fully}. 

The network employs a progressive feature extraction strategy. Initial convolutional layers capture low-level spatial features and fundamental image patterns. MAN modules in deeper encoding stages process high-level semantic representations through dual-branch architectures. PAGF modules replace traditional skip connections \cite{ronneberger2015u} to enable intelligent feature fusion while preserving critical spatial-channel information. These architectures combine KAN blocks with multi-scale attention processing \cite{vaswani2017attention} and adaptive fusion mechanisms.

\subsection{Multi-Scale Adaptive Network Enhancement}

\begin{figure}[htbp]
    \centering
    \includegraphics[width=\columnwidth]{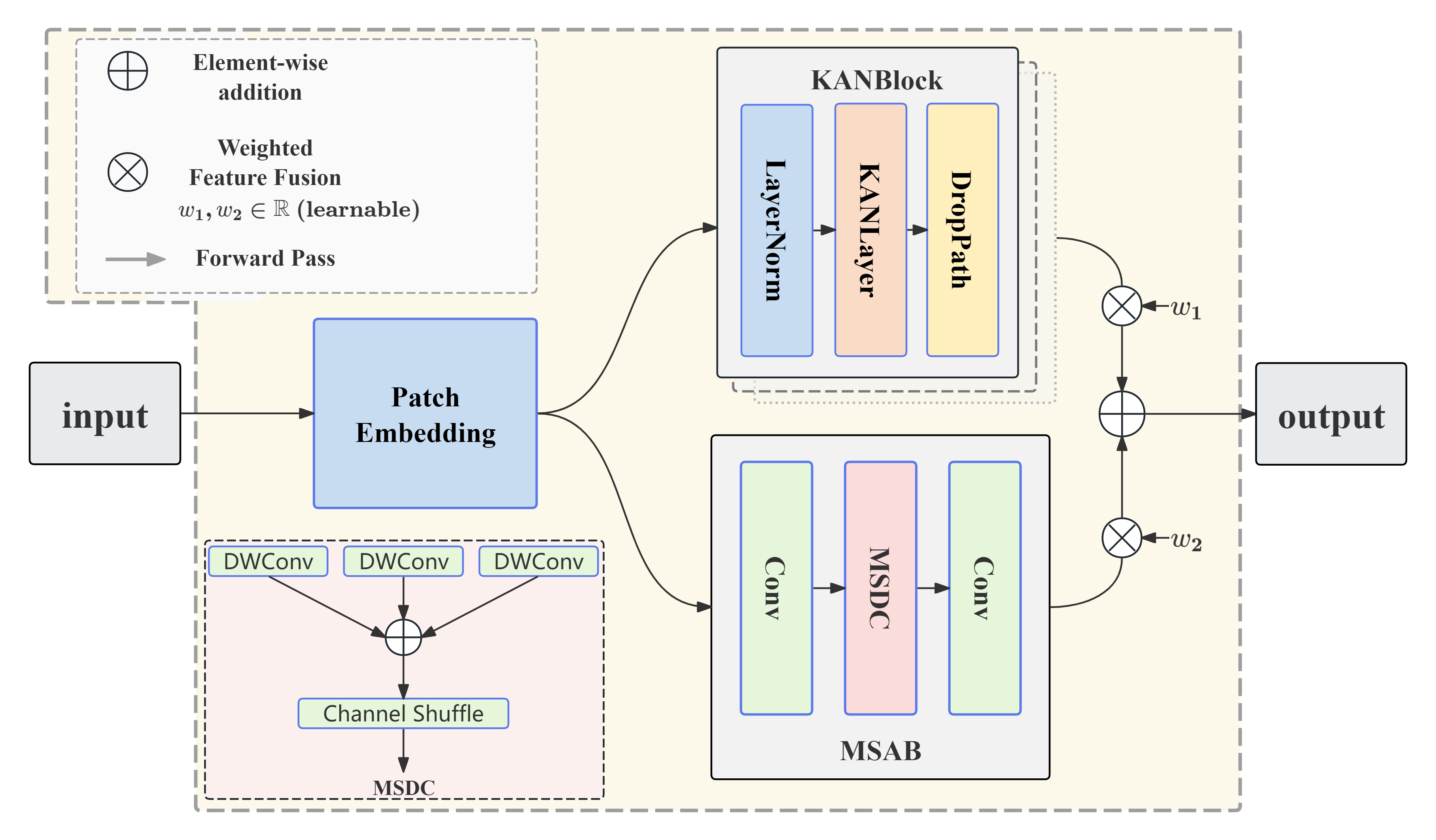}
    \caption{Architecture of MAN module. The structure employs dual-branch processing with KAN blocks for learnable activation-based feature processing and MSAB blocks for multi-scale attention feature extraction, combined through adaptive fusion mechanisms.}
    \label{fig:MAN}
\end{figure}

To enhance the expressive ability and multi-scale extraction capability of the model, we propose MAN modules. As illustrated in Figure \ref{fig:MAN}, MAN combines KAN-based feature processing with Multi-Scale Attention Blocks (MSAB) to enable comprehensive multi-scale feature extraction \cite{vaswani2017attention}. 

The computational flow begins by feeding the input into a Patch Embedding module \cite{dosovitskiy2020image}, which converts the raw data into a sequence of embedded patches. Subsequently, these patches are fed into a dual-branch architecture. The primary branch consists of KANBlocks, which refine features by sequentially applying Layer Normalization, a KAN Layer, and DropPath regularization \cite{huang2016deep}. In parallel, the secondary branch, known as the MSAB, is dedicated to extracting multi-scale features through a cascade of convolutional, batch normalization, and ReLU layers.

The module processes an input feature map, denoted as $\mathbf{X}_{\text{embed}} \in \mathbb{R}^{H_{embed} \times W_{embed} \times C}$, through a dual-branch architecture \cite{he2016deep}. This parallel design allows the network to simultaneously capture complex non-linear relationships and diverse spatial patterns.

The primary branch leverages a standard KANBlock to act as an efficient and interpretable feature embedder. It processes the input features to capture intricate and high-frequency signal details, leveraging the superior function approximation capabilities of KAN \cite{liu2024kan}. The output of this branch is formulated as:
\begin{equation}
    \mathbf{F}_{\text{kan}} = \text{KANBlock}(\mathbf{X}_{\text{embed}})
    \label{eq:kan_branch}
\end{equation}
where $\mathbf{F}_{\text{kan}}$ represents the feature map refined by the KAN layer.

In parallel, the secondary branch employs a Multi-Scale Attention Block (MSAB) to explicitly extract spatial features at various scales \cite{zhao2017pyramid}. As shown in the detailed view of MSAB in Figure~\ref{fig:MAN}, this block consists of a sequence of convolutional layers, including a multi-scale depthwise convolution (MSDC) unit \cite{howard2017mobilenets}, designed to capture a wide range of contextual information. The output of this attention-focused branch is given by:
\begin{equation}
    \mathbf{F}_{\text{msab}} = \text{MSAB}(\mathbf{X}_{\text{embed}})
    \label{eq:msab_branch}
\end{equation}
where $\mathbf{F}_{\text{msab}}$ contains multi-scale spatial context.

We introduce learnable scalar weights, $w_1$ and $w_2$, to dynamically balance the contributions of the multi-scale features from the MSAB \cite{hu2018squeeze}. The final output of the MAN module, $\mathbf{F}_{\text{man}}$, is computed as:
\begin{equation}
    \mathbf{F}_{\text{man}} = w_{1} \otimes \mathbf{F}_{msab} + w_{2} \otimes \mathbf{F}_{kan} 
    \label{eq:man_fusion}
\end{equation}
where $w_{1}$ and $w_{2}$ represent learnable fusion weights \cite{wang2018non} that adaptively balance multi-scale attention features and KAN-processed representations.

\subsection{Progressive Attention-Guided Feature Fusion}
The PAGF module is an advanced feature fusion mechanism motivated by the need to alleviate the semantic gap of indiscriminate fusion in traditional U-Net skip connections \cite{ronneberger2015u}, which can introduce noise and irrelevant information. PAGF aims to replace simple feature concatenation with a learnable, adaptive mechanism \cite{oktay2018attention} that can intelligently arbitrate and fuse features from the encoder and decoder paths to improve segmentation accuracy.

\par\medskip\noindent
The design of PAGF is both concise and effective. Given up-sampled decoder features $X_d$ and detailed encoder features $X_e$, the module first concatenates them for contextual information \cite{long2015fully}. It then concurrently generates a comprehensive attention map $A$ to focus on salient feature regions and a gating map $G$ to dynamically determine the fusion ratio at each spatial location \cite{chen2018encoder}. The final fusion process is governed by the equation:
\begin{equation}
    F_{fused} = G \otimes (X_d \otimes A) + (1 - G) \otimes (X_e \otimes A)
\end{equation}
This equation performs a weighted sum of the attention-enhanced features, guided by the data-driven gating signal $G$, to achieve a highly flexible fusion \cite{woo2018cbam}. The resulting features then undergo a final refinement step through a convolutional layer.

\begin{figure}[htbp]
    \centering
    \includegraphics[width=\columnwidth]{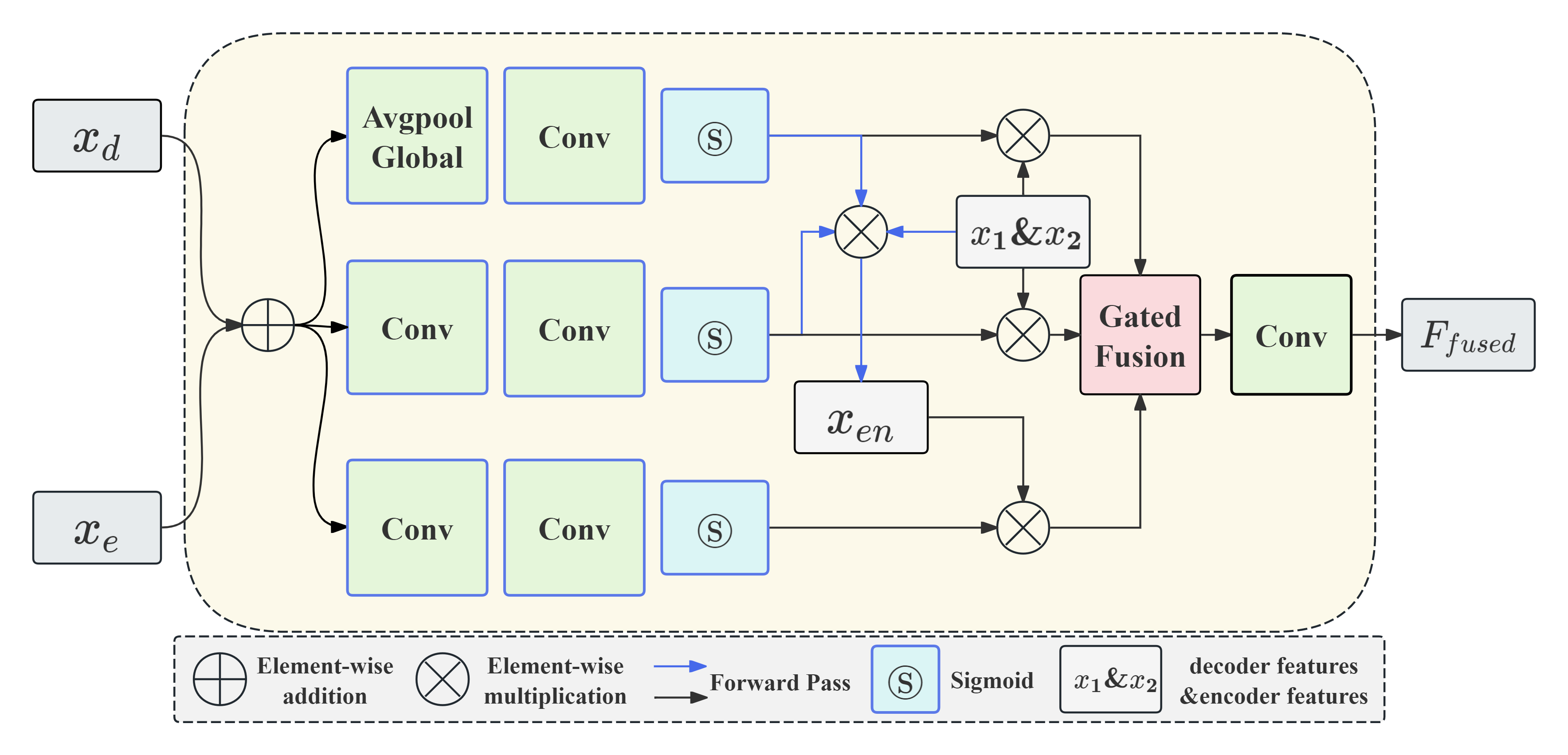}
    \caption{Architecture of PAGF module. The structure employs dual-path attention with channel and spatial mechanisms, followed by gated fusion for intelligent encoder-decoder feature combination.}
    \label{fig:PAGF}
\end{figure}

\subsection{Training Strategy and Implementation Details}
To optimize U-MAN, we employ a combined loss function optimizing both region-based and boundary-based segmentation accuracy \cite{milletari2016v}:
\begin{equation}
\mathcal{L}_{total} = \lambda_{dice} \times \mathcal{L}_{dice} + \lambda_{bce} \times \mathcal{L}_{bce}
\end{equation}
where $\mathcal{L}_{dice}$ represents Dice loss for region overlap optimization and $\mathcal{L}_{bce}$ denotes binary cross-entropy loss for pixel-wise classification \cite{de2005tutorial}.

During inference, U-MAN processes input images through MAN-enhanced encoder pathways followed by PAGF-guided decoder reconstruction.

\section{EXPERIMENTS}
\subsection{Experimental Settings}
\subsubsection{Datasets}
We evaluate U-MAN on three widely used medical image segmentation datasets. These datasets cover different imaging modalities and anatomical structures. The CVC-ClinicDB dataset \cite{bernal2015wm} contains 612 colonoscopy images with polyp annotations for endoscopic polyp segmentation. The GLAS dataset \cite{sirinukunwattana2017gland} consists of 165 histopathological images from the Warwick-QU dataset for glandular structure segmentation. The BUSI dataset \cite{al2020dataset} includes 647 breast ultrasound images with lesion masks. These datasets provide challenging cases with varying tissue characteristics and boundary definitions. 

\subsubsection{Implementation Details}
All experiments are conducted using PyTorch framework \cite{paszke2019pytorch} on NVIDIA GeForce RTX 3090 GPU. The network employs embedding dimensions of [32, 64, 256, 320, 512] and MAN module depths of [3, 3, 3]. Training uses differentiated learning rates for different components. Data augmentation includes random rotation, flipping, and normalization following standard medical image segmentation protocols \cite{shorten2019survey}. All datasets are resized to 256×256 pixels and split into 80\% training and 20\% validation sets using random seeds for reproducibility.

\begin{table*}[!htbp]
    \centering
    \caption{Quantitative comparison of different methods on three datasets. Best results are highlighted in bold.}
    \label{tab:quantitative_results}
    \resizebox{0.75\textwidth}{!}{%
    \tiny
    \renewcommand{\arraystretch}{0.7}%
    \begin{tabular}{l|cc|cc|cc}
        \hline
        \multirow{2}{*}{Method} & \multicolumn{2}{c|}{BUSI} & \multicolumn{2}{c|}{GLAS} & \multicolumn{2}{c}{CVC-ClinicDB} \\
        & IoU(\%) & F1(\%) & IoU(\%) & F1(\%) & IoU(\%) & F1(\%) \\
        \hline
        U-Net \cite{ronneberger2015u} & 57.22 & 71.91 & 86.66 & 92.79 & 83.79 & 91.06 \\
        Attention U-Net \cite{oktay2018attention} & 55.18 & 70.22 & 86.84 & 92.89 & 84.52 & 91.46 \\
        U-Net++ \cite{zhou2018unet} & 57.41 & 72.11 & 87.07 & 92.96 & 84.61 & 91.53 \\
        U-NeXt \cite{valanarasu2022unext} & 59.06 & 73.08 & 84.50 & 91.55 & 74.83 & 85.56 \\
        Rolling-UNet \cite{zhang2023rolling} & 61.00 & 74.67 & 86.42 & 92.63 & 82.87 & 90.48 \\
        U-Mamba \cite{ma2024u} & 61.81 & 75.55 & 87.01 & 93.02 & 84.79 & 91.63 \\
        U-KAN \cite{li2024u} & 64.99 & 78.55 & 87.74 & 93.47 & 82.69 & 90.24 \\
        \hline
        \textbf{U-MAN (Ours)} & \textbf{66.12} & \textbf{79.12} & \textbf{89.07} & \textbf{94.21} & \textbf{86.14} & \textbf{92.47} \\
        \hline
        \rowcolor{green!10} 
        \textbf{improve} & \textbf{1.13\%} & \textbf{0.57\%} & \textbf{1.33\%} & \textbf{0.74\%} & \textbf{1.35\%} & \textbf{0.84\%} \\
        \hline
    \end{tabular}%
    }
\end{table*}

\subsection{Main Results}

\begin{figure}[htbp]
\centering
\includegraphics[width=\columnwidth]{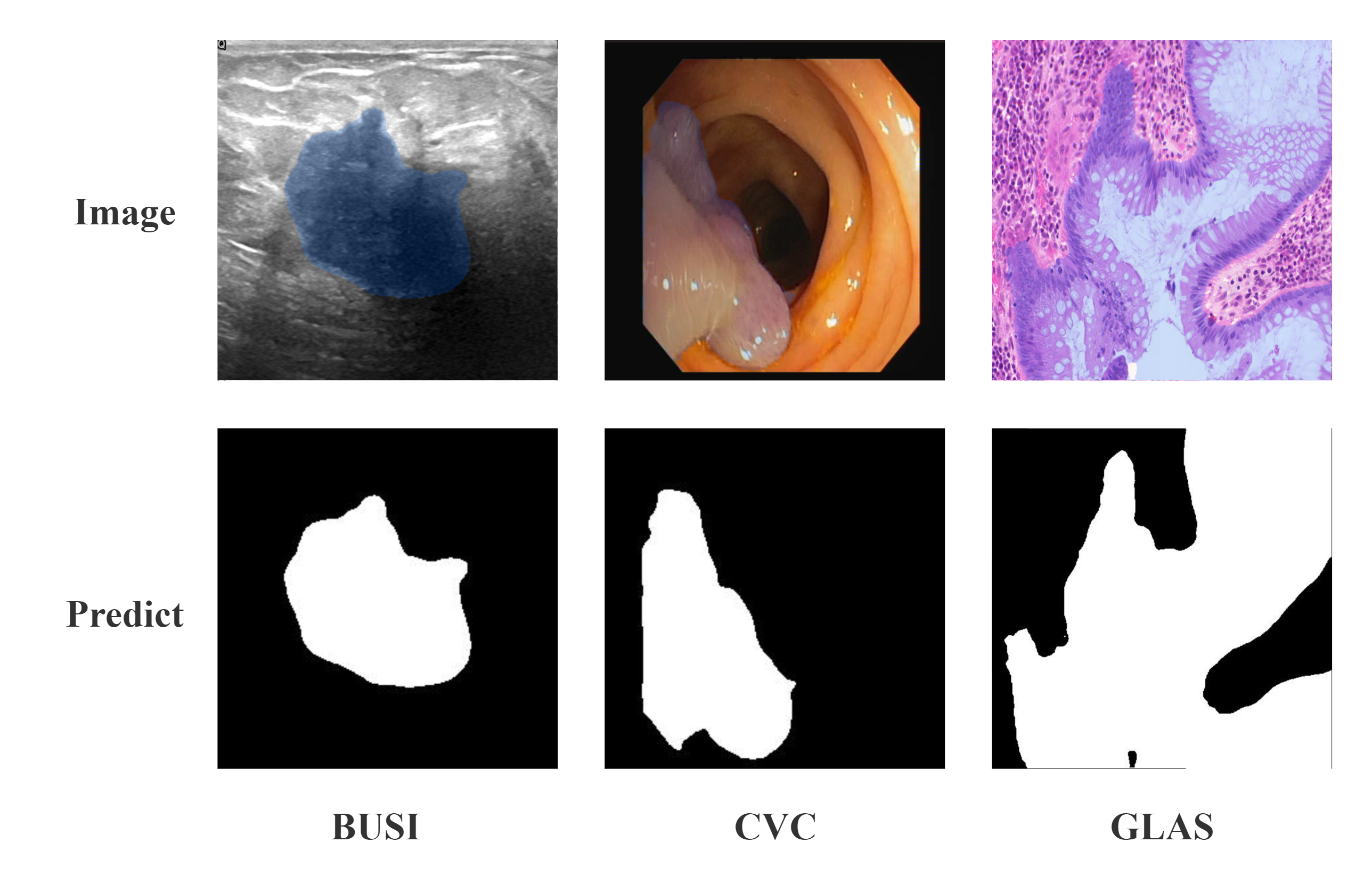}
\vspace{-0.35cm}
\caption{Qualitative segmentation results of U-MAN across three medical imaging datasets. The top row shows original images from BUSI, CVC-ClinicDB, and GLAS. The bottom row presents corresponding binary segmentation masks generated by our method, demonstrating accurate boundary delineation across diverse medical imaging modalities.} 
\label{fig:segmentation_results}

\end{figure}

Table~\ref{tab:quantitative_results} presents quantitative comparisons with state-of-the-art methods across all three datasets using standard evaluation metrics. As illustrated in Figure \ref{fig:segmentation_results}, U-MAN consistently achieves superior performance, demonstrating the effectiveness of our PAGF approach combined with MAN.

U-MAN achieves the best performance across all datasets. On BUSI \cite{al2020dataset}, we attain 66.12\% IoU and 79.12\% F1-score, with 1.13\% IoU improvement over U-KAN \cite{li2024u}. For GLAS \cite{sirinukunwattana2017gland}, our method reaches 89.07\% IoU and 94.21\% F1-score, outperforming baselines by 1.33\% IoU. On CVC-ClinicDB \cite{bernal2015wm}, we achieve 86.14\% IoU and 92.47\% F1-score, representing substantial 3.45\% IoU improvement over U-KAN.

The consistent performance gains validate the generalizability of our PAGF mechanism and MAN architecture. Compared to recent advanced methods like U-Mamba \cite{ma2024u} (84.79\% IoU on CVC-ClinicDB), our approach demonstrates superior segmentation capabilities across diverse medical imaging modalities.

\subsection{Ablation Study}

We conduct systematic ablation experiments on the CVC-ClinicDB dataset \cite{bernal2015wm} to validate the effectiveness of each proposed component. The study is organized into three hierarchical levels to comprehensively analyze our design choices.

\subsubsection{Ablation Study on Overall Components}

\vspace{-10pt}

\begin{table}[htbp]
\centering
\caption{Overall component ablation and computational cost analysis on CVC-ClinicDB.}
\label{tab:overall_ablation}
\setlength{\tabcolsep}{3pt}
\begin{tabular}{l|cc}
\hline
\textbf{Configuration} & \textbf{IoU (\%)} & \textbf{F1 (\%)} \\
\hline
U-KAN (Baseline) \cite{li2024u} & 82.69 & 90.24 \\
U-MAN (Full) & \textbf{86.14} & \textbf{92.47} \\
w/o MAN Module & 83.82 & 90.95 \\
w/o PAGF Module & 84.91 & 91.89 \\
\hline
\end{tabular}
\end{table}

U-MAN achieves substantial 3.45\% IoU improvement over U-KAN baseline \cite{li2024u}, demonstrating the effectiveness of our proposed architecture. MAN module contributes 2.32\% IoU gain (86.14\% vs 83.82\%), highlighting its critical role in multi-scale feature processing. PAGF module adds 1.23\% improvement (86.14\% vs 84.91\%) through intelligent skip connections.The combined effect (3.45\%) is slightly less than the sum of individual contributions (2.32\% + 1.23\% = 3.55\%), suggesting minor overlap in the functionalities of the two modules while still demonstrating their collective necessity for optimal performance.

\subsubsection{Ablation Study on MAN Module}

\begin{table}[htbp]
\centering
\caption{MAN module component ablation results on CVC-ClinicDB.}
\label{tab:man_ablation}
\begin{tabular}{l|cc}
\hline
\textbf{Configuration} & \textbf{IoU (\%)} & \textbf{F1 (\%)} \\
\hline
w/o MSAB (KAN only) & 85.67 & 92.18 \\
Single-scale (3×3) & 85.49 & 92.05 \\
Multi-scale (1×1,3×3,5×5) & \textbf{86.14} & \textbf{92.47} \\
1-layer KAN & 84.78 & 91.73 \\
3-layer KAN (Optimal) & \textbf{86.14} & \textbf{92.47} \\
5-layer KAN & 85.82 & 92.29 \\
\hline
\end{tabular}
\end{table}

MSAB component contributes 0.47\% IoU improvement (86.14\% vs 85.67\%) over pure KAN processing \cite{liu2024kan}, validating the importance of multi-scale attention mechanisms \cite{zhang2019attention}. The optimal \{1×1, 3×3, 5×5\} kernel configuration outperforms single-scale by 0.65\% (86.14\% vs 85.49\%), effectively capturing both fine details and global context. For KAN depth analysis, 3-layer configuration achieves optimal performance with 1.36\% improvement over 1-layer (86.14\% vs 84.78\%), while 5-layer networks show overfitting with 0.32\% degradation (85.82\% vs 86.14\%), confirming our architectural design choice.

\vspace{-10pt}

\subsubsection{Ablation Study on PAGF}

\vspace{-10pt}

\begin{table}[htbp]
\centering
\caption{PAGF module component ablation analysis on CVC-ClinicDB.}
\label{tab:pagf_ablation}
\begin{tabular}{l|cc}
\hline
\textbf{Configuration} & \textbf{IoU (\%)} & \textbf{F1 (\%)} \\
\hline
Simple Skip Connection & 83.82 & 90.95 \\
w/o Channel Attention & 85.21 & 92.08 \\
w/o Spatial Attention & 85.58 & 92.25 \\
w/o Gating Mechanism & 85.89 & 92.36 \\
Element-wise Addition & 84.67 & 91.59 \\
Full PAGF & \textbf{86.14} & \textbf{92.47} \\
\hline
\end{tabular}
\end{table}

Channel attention demonstrates superior contribution (0.93\% improvement: 86.14\% vs 85.21\%) compared to spatial attention (0.56\% improvement: 86.14\% vs 85.58\%), suggesting feature channel recalibration is more critical for medical segmentation \cite{hu2018squeeze}. The gating mechanism \cite{dauphin2017language} provides additional 0.25\% improvement (86.14\% vs 85.89\%) for adaptive information flow. Our attention gate mechanism significantly outperforms element-wise addition by 1.47\% IoU (86.14\% vs 84.67\%), demonstrating the effectiveness of learnable adaptive fusion \cite{oktay2018attention} over simple arithmetic operations.

\section{CONCLUSION}

In this paper, we propose U-MAN, a novel network for medical image segmentation. This method is designed to address two core limitations of conventional U-Net architectures: (1) simple skip connections that ignore the semantic gap between encoder and decoder features, and (2) a lack of adaptive multi-scale processing capabilities in deep layers. To achieve this, U-MAN enhances the emerging U-KAN architecture with two key innovations: the Progressive Attention-Guided Feature Fusion (PAGF) module, which replaces simple skip connections with an attention mechanism for feature fusion; and the Multi-scale Adaptive kAN (MAN) module, which empowers the network to adaptively process features at multiple scales. Experiments demonstrate the superiority of U-MAN, which achieves IoU improvements of 1.13\%, 1.33\%, and 3.45\% over the U-KAN baseline on the BUSI, GLAS, and CVC-ClinicDB datasets, respectively. These results validate that U-MAN produces segmentations with finer details and more precise boundaries across diverse medical imaging modalities.

    \bibliographystyle{IEEEtran}
    \bibliography{refs}
    \balance

\end{document}